\documentclass[orivec]{llncs}
\usepackage[T1]{fontenc}
\usepackage{graphicx}
\usepackage{amsmath}
\usepackage{booktabs}
\usepackage{multirow}
\usepackage{xcolor}
\usepackage{enumitem}
\usepackage{hyperref}
\usepackage{cite}
\usepackage{float}
\usepackage{subfigure}    
\usepackage{algorithm}
\usepackage{algorithmic}
\usepackage{wrapfig}
\usepackage{geometry}

\begin{document}

\title{MATEX: Multi-scale Attention and Text-guided Explainability of Medical Vision-Language Models}

\author{Muhammad Imran$^1$ \and Chi Lee$^2$ \and Yugyung Lee$^1$}
\authorrunning{Imran et al.}
\institute{
$^1$Computer Science, School of Science and Engineering,\\
$^2$Division of Pharmacology and Pharmaceutical Sciences, School of Pharmacy,\\
University of Missouri - Kansas City, USA \\
\email{\{mi3dr, leech, leeyu\}@umkc.edu}
}

\maketitle
\begin{abstract}
We introduce MATEX (Multi-scale Attention and Text-guided Explainability), a novel framework that advances interpretability in medical vision-language models by incorporating anatomically informed spatial reasoning. MATEX synergistically combines multi-layer attention rollout, text-guided spatial priors, and layer consistency analysis to produce precise, stable, and clinically meaningful gradient attribution maps. By addressing key limitations of prior methods—such as spatial imprecision, lack of anatomical grounding, and limited attention granularity—MATEX enables more faithful and interpretable model explanations. Evaluated on the MS-CXR dataset, MATEX outperforms the state-of-the-art M2IB approach in both spatial precision and alignment with expert-annotated findings. These results highlight MATEX’s potential to enhance trust and transparency in radiological AI applications.

\end{abstract}

\section{Introduction}

Recent advances in Vision-Language Models (VLMs) such as CLIP~\cite{radford2021learning}, BioViL~\cite{boecking2022making}, and GLoRIA~\cite{huang2021gloria} have demonstrated impressive capabilities in medical imaging tasks, including chest X-ray classification~\cite{boecking2022making}. By bridging visual features with clinical text, these models offer promise for automated diagnosis, triage, and decision support. However, their growing utility is hindered by a fundamental limitation: a lack of interpretability. Without transparent, trustworthy explanations, these models remain ill-suited for high-stakes clinical environments, where accountability, safety, and human oversight are paramount.

In clinical settings, interpretability is not a luxury—it is a prerequisite for adoption. Physicians must be able to trace AI predictions back to meaningful visual evidence, assess whether those predictions align with known pathological findings, and understand when and why models may fail. Furthermore, developers rely on interpretability to uncover latent biases, assess failure modes, and refine model behavior~\cite{wang2023visual}. The ability to generate attribution maps that are both spatially precise and clinically grounded is essential for building trust in AI-assisted diagnostics.

Conventional gradient-based interpretability techniques—such as Saliency Maps~\cite{simonyan2014deep}, Grad-CAM~\cite{selvaraju2016grad}, and Integrated Gradients~\cite{sundararajan2017axiomatic}—offer limited insight into model behavior. These methods often suffer from low spatial resolution, high noise sensitivity, and lack of anatomical correspondence. Transformer-based VLMs, while powerful, introduce additional complexity through deep self-attention layers, making it difficult to trace meaningful signal propagation~\cite{abnar2020quantifying,chefer2021generic}.

To overcome these challenges, recent approaches such as RISE~\cite{rise} and M2IB~\cite{m2ib} incorporate perturbation-based analysis and information bottlenecks to enhance attribution quality. Despite their progress, these methods still fall short in key areas: they lack anatomical priors, struggle with layer-wise attention inconsistency, and rarely incorporate language-grounded spatial guidance—an essential element for multimodal medical understanding.

\begin{wrapfigure}{r}{0.48\textwidth}
\vspace{-10pt}
\centering
\includegraphics[width=0.46\textwidth]{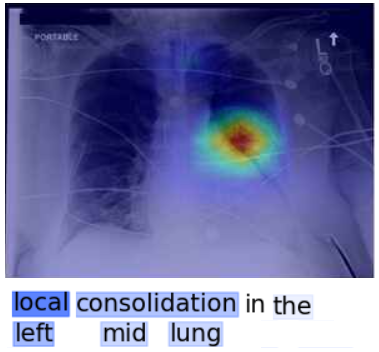}
\vspace{-10pt}
\caption{\textbf{MATEX} Text-aware and anatomically grounded explanations produced by MATEX.
Heatmaps localize lung cancer regions in the input image, while highlighted caption keywords indicate the clinical concepts driving the prediction.}
\label{fig:main_marts}
\vspace{-10pt}
\end{wrapfigure}

We propose \textbf{MATEX} (Multi-scale Attention and Text-guided Explainability), a novel framework designed to bridge this gap. MATEX generates anatomically faithful and text-aware attribution maps by leveraging three core innovations: (1) a multi-scale attention rollout strategy to capture signal flow across transformer layers, (2) clinically informed spatial priors derived from radiological text, and (3) a layer consistency mechanism to filter out unstable gradients. In this sense, MATEX serves as an interpretability framework for large vision–language models in medical imaging, analogous to the role of M$^2$IB in explaining multimodal reasoning.
As shown in Figure~\ref{fig:main_marts}, MATEX provides joint visual–textual explanations by localizing lung cancer regions through anatomically consistent heatmaps while simultaneously highlighting the caption keywords that guide these attributions. This alignment between image evidence and clinical language enables transparent inspection of LVLM decision-making, closely reflecting radiologist-style reasoning and facilitating trustworthy deployment in medical settings.

Our contributions are threefold:
\begin{itemize}[leftmargin=*]
    \item \textbf{Multi-Scale Attention Aggregation:} We extend attention rollout to fuse spatial features across transformer layers, improving attribution granularity and robustness.
    \item \textbf{Text-Guided Spatial Priors:} We incorporate structured clinical language to anchor gradient attribution maps around anatomically relevant regions.
    \item \textbf{Layer Consistency Filtering:} We propose a cross-layer consistency scheme to suppress noisy or unstable signals, leading to more reliable and interpretable visual explanations.
\end{itemize}

Extensive evaluations on the MS-CXR benchmark~\cite{boecking2022making} show that MATEX surpasses existing interpretability methods—including M2IB~\cite{m2ib} and Chefer et al.~\cite{chefer2021generic}—in both quantitative metrics (confidence change, ROAR+) and qualitative alignment with expert annotations. Our results highlight the importance of integrating textual and architectural priors for clinically trustworthy AI explanations.

\section{Related Work}

\textbf{Medical Vision-Language Models.}  
Vision-Language Models (VLMs) are increasingly used in medical imaging to align visual features with clinical text. CLIP~\cite{radford2021learning} catalyzed this trend, inspiring domain-specific adaptations. BioViL~\cite{boecking2022making} fine-tuned vision-language alignment on chest X-rays, while GLoRIA~\cite{huang2021gloria} enhanced spatial grounding of radiological phrases, and Zhou et al.~\cite{zhou2021visual} proposed attentive semantic consistency for report generation. Yet, most models remain black boxes, offering limited insight into how clinical descriptions guide visual predictions, thus hindering clinical trust and adoption. Recent surveys further highlight these limitations, noting that while medical VLMs show promise across retrieval, report generation, and grounding tasks, issues of interpretability, reliability, and trust remain largely unresolved and inconsistently evaluated across datasets and clinical settings~\cite{imran2026vlmsurvey}. Broader reviews also highlight these concerns across modalities, noting persistent challenges in fairness, transparency, and ethical trustworthiness in multimodal AI systems~\cite{saleh2025trustworthy, wang2025vlmmed, lin2025survey}.

\textbf{Trust and Reliability in Vision-Language Models.}  
Beyond interpretability, recent work has emphasized the importance of estimating when VLM predictions should be trusted. Imran et al.~\cite{imran2026trustvlm} show that vision-language models frequently exhibit overconfidence in spatial reasoning tasks, and propose confidence estimation mechanisms to distinguish reliable from unreliable predictions. However, such approaches focus on prediction-level trust rather than providing localized, anatomically grounded explanations—leaving open the question of how trust signals align with interpretable visual evidence.

\textbf{Gradient-Based Attribution Methods.}  
Classic attribution methods like Saliency Maps~\cite{simonyan2014deep}, Grad-CAM~\cite{selvaraju2016grad}, and Integrated Gradients~\cite{sundararajan2017axiomatic} highlight pixel importance but often suffer from noise and poor localization. Extensions like CLIP-IG~\cite{zhao2024gradient} adapt these for multimodal tasks but still lack anatomical grounding. Most approaches also neglect clinical context—e.g., anatomical regions or directional cues—which are critical in radiology.

\textbf{Attention-Based Interpretability.}  
Attention-based methods visualize internal transformer dynamics. Abnar and Zuidema~\cite{abnar2020quantifying} introduced attention rollout, while Chefer et al.~\cite{chefer2021generic} extended attribution to encoder-decoder models. Though model-intrinsic, such maps are often diffuse and not clinically meaningful. M2IB~\cite{m2ib} addressed this with information bottlenecks, yet without text or anatomical priors, its interpretability remains limited.

\textbf{Interpretability in Medical Imaging.}  
Interpretability in medical imaging demands anatomical accuracy and alignment with clinical language. Visual bottlenecks~\cite{wang2023visual} and contrastive methods~\cite{ketabi2025multimodal} have improved saliency, but often ignore linguistic cues such as ``right upper lobe'' or ``posterior region.'' Recent multimodal approaches attempt to bridge this gap. Imran and Lee~\cite{imran2025multi} propose a multi-modal interpretability framework that combines visual and textual signals to improve localization fidelity in vision--language models, demonstrating more coherent alignment between attention and semantic regions. However, such methods primarily focus on localization quality and do not explicitly address whether the resulting explanations correspond to reliable or trustworthy model predictions. Moreover, most evaluations remain qualitative, as noted in iMIMIC’s call for structured, domain-grounded interpretability benchmarks.

These gaps across model design, attribution, and evaluation motivate our proposed \textbf{MATEX} framework, which integrates gradient flow, attention rollout, and text-informed spatial priors to generate anatomically and semantically faithful explanations.

\section{Methods}

\subsection{MATEX Framework Overview}

\begin{figure}[t]
\centering
\includegraphics[width=\textwidth]{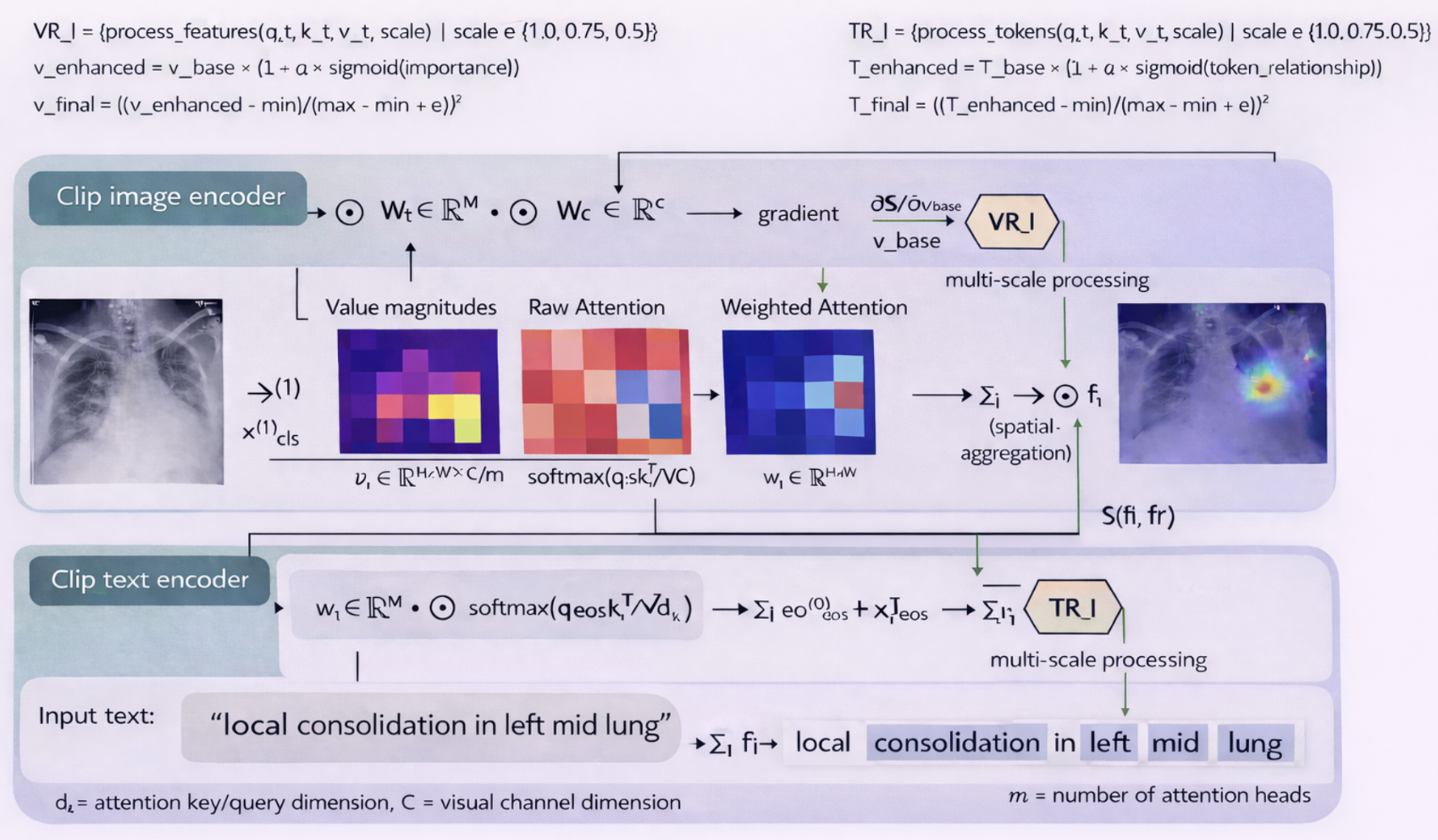}
\caption{\textbf{MATEX architecture.} \textbf{Top:} the CLIP image encoder embeds a chest X-ray and computes value magnitudes, raw attention, and weighted spatial attention; a similarity score $S(\mathbf{f}_i,\mathbf{f}_r)$ provides gradients $\partial S/\partial \mathbf{v}_{\text{base}}$ that drive the visual relevance module (VR\_I) via multi-scale enhancement to produce a spatial explanation map. \textbf{Bottom:} the CLIP text encoder embeds the input report phrase (e.g., ``local consolidation in left mid lung'') and applies token attention (scaled by $\sqrt{d_k}$) to obtain token relevance (TR\_I) through multi-scale processing, yielding token-level attributions and anatomical cues. The final explanation combines VR\_I and TR\_I to align salient image regions with clinically meaningful tokens. Here $d_k$ is the attention key/query dimension, $C$ is the visual channel dimension, and $m$ is the number of attention heads.}
\label{fig:architecture}
\end{figure}

\textbf{MATEX} (Multi-Scale Attention and Text-guided Explainability) is an interpretable framework for medical vision-language models (VLMs), developed to generate clinically grounded and anatomically faithful explanations. The system combines hierarchical transformer attention, gradient attribution, and text-guided spatial priors, as illustrated in Figure~\ref{fig:architecture}.

The architecture consists of five interconnected components:

\begin{itemize}[leftmargin=*, noitemsep]
    \item \textbf{Visual Attention Maps (vmaps):} Derived from the image encoder using value weights, raw attention, and gradient backpropagation. Multi-scale rollout is enhanced via the VR\_I module, producing anatomically relevant saliency heatmaps.
    
    \item \textbf{Textual Attention Maps (tmaps):} From the text encoder, token embeddings are weighted via attention mechanisms and processed by the TR\_I module to emphasize clinically important phrases. Token relevance scores are aligned with regions in the image.
    
    \item \textbf{Gradient Attribution:} Gradients of the image-text similarity score $S(f_i, f_r)$ are computed with respect to both the image features $o^{(0)}_{\text{eos}}$ and the text features $o^{(0)}_{\text{cls}}$. These gradients highlight the sensitivity of specific patches and tokens to the alignment decision.
    
    \item \textbf{Text-Guided Spatial Priors:} Anatomical references in text (e.g., “left lower lung”) are parsed and converted into 2D spatial priors using Gaussian region maps. These soft priors guide visual explanation toward clinically meaningful locations.
    
   \item \textbf{Multi-Scale Fusion:} All attribution maps—gradient, attention rollout, consistency, and text priors—are combined via a tunable fusion scheme to produce the final explanation map $A_{\text{MATEX}}$.

\end{itemize}
\subsection{Multi-Scale Attention Rollout and Consistency}

As visualized in Figure~\ref{fig:architecture}, MATEX begins by processing a chest X-ray image through a transformer-based visual encoder (e.g., CLIP ViT). This encoder produces attention maps across multiple layers, capturing hierarchical visual semantics: early layers attend to low-level structural patterns (e.g., lung borders), while deeper layers highlight more abstract clinical features (e.g., lesions or opacities).

To leverage this multi-level information, MATEX performs \textit{multi-scale attention rollout} from the [CLS] token to image patches. This operation traces how attention flows through the network layers, producing a comprehensive attention map $A_{\text{flow}}$:

\begin{equation}
A_{\text{flow}} = \sum_{l=1}^{L} w_l \cdot \text{Attention}^{(l)}_{[\text{CLS}] \rightarrow \text{patches}}, \quad 
w_l = \frac{\exp(\tau l)}{\sum_{k=1}^{L} \exp(\tau k)}
\end{equation}

Here, $L$ is the total number of transformer layers, and the temperature parameter $\tau$ controls the weighting scheme, placing more emphasis on deeper layers when identifying pathology-related semantics.

However, attention scores can vary significantly between layers, particularly in ambiguous or noisy regions. To mitigate this, MATEX introduces a \textit{consistency scoring module} (shown in the figure as a blue consistency block following the attention rollout stream). For each spatial patch $i$, we compute a reliability score $C_i$ based on the variance of attention values across layers:

\begin{equation}
C_i = \frac{1}{1 + \sigma_l(\text{Attention}^{(l)}_{[\text{CLS}] \rightarrow i})}
\end{equation}

This scoring function penalizes inconsistent patches—those where attention fluctuates across layers—and highlights stable, semantically grounded regions. These scores are later used to modulate the gradient attribution map $A_{\text{grad}}$, as illustrated by the fusion pathway in the figure where $C \odot A_{\text{grad}}$ is computed before entering the fusion module.

\textbf{Clinical Relevance.} For example, in radiographs where both the left and right lungs exhibit subtle opacities, attention across layers may drift. The consistency score $C_i$ suppresses such instability, helping MATEX focus its explanation on robust findings like well-formed consolidations, especially when aligned with gradient or text-derived signals.

By explicitly modeling both the hierarchical attention flow and its cross-layer consistency, this component ensures that explanations are not only attentive to relevant features but also stable across model depths, essential for clinical trustworthiness and reproducibility.

\subsection{Text-Guided Spatial Priors}

As illustrated in the purple pathway in Figure~\ref{fig:architecture}, MATEX processes the clinical report alongside the image to derive anatomical guidance. The text branch begins with a transformer-based encoder (e.g., CLIP's text encoder), which extracts token-level representations of the report. To transform these linguistic features into spatially relevant guidance, MATEX incorporates a \textbf{rule-based parser} that identifies anatomical references, such as "left lower lobe" or "right apex".

These phrases are mapped to a set of anatomical regions $R$, each associated with a corresponding spatial zone defined by coordinate ranges in normalized $(x, y)$ image coordinates. For example:
\begin{itemize}[noitemsep]
    \item "right apex" $\rightarrow$ upper-right lung region: $x \in [0.5, 1.0], y \in [0.0, 0.4]$
    \item "left base" $\rightarrow$ lower-left lung region: $x \in [0.0, 0.5], y \in [0.6, 1.0]$
\end{itemize}

For each region $r \in R$, we define smooth coordinate range weights that create soft transitions within the specified anatomical boundaries. These weights are computed using clamped linear interpolation to ensure smooth falloff at region boundaries:

\begin{align}
w_x^{(r)} &= \text{clamp}\left(\frac{x - x_{\min}}{x_{\max} - x_{\min}}, 0, 1\right) \cdot \text{clamp}\left(\frac{x_{\max} - x}{x_{\max} - x_{\min}}, 0, 1\right) \\
w_y^{(r)} &= \text{clamp}\left(\frac{y - y_{\min}}{y_{\max} - y_{\min}}, 0, 1\right) \cdot \text{clamp}\left(\frac{y_{\max} - y}{y_{\max} - y_{\min}}, 0, 1\right)
\end{align}

The final spatial prior combines contributions from all detected regions and applies max-normalization for stability:

\begin{equation}
M(x,y) = 1 + (\lambda_s - 1) \cdot \text{normalize}\left(\sum_{r \in R} w_x^{(r)} \cdot w_y^{(r)}\right)
\end{equation}

where $\text{normalize}(\cdot)$ performs min-max normalization: $\text{normalize}(z) = \frac{z - \min(z)}{\max(z) - \min(z)}$ to ensure consistent scaling across different numbers of detected regions.

The parameter $\lambda_s \geq 1$ controls the strength of the text-derived spatial bias: higher values increase the influence of text priors on the final attribution map. When no anatomical text is detected, $M(x,y) = 1$, and the prior becomes neutral.

\textbf{Integration and Role.} As shown in the fusion block of Figure~\ref{fig:architecture}, the spatial prior $M(x,y)$ modulates the gradient map $A_{\text{grad}}$ through an elementwise product $M \odot A_{\text{grad}}$, encouraging explanations to focus on medically relevant regions inferred from the text.

\textbf{Clinical Motivation.} For instance, if a radiology report references "opacity in the left lower lung", but the model's raw attribution highlights unrelated areas (e.g., mediastinum), the spatial prior will shift the attribution map toward the correct anatomical zone, improving interpretability and alignment with expert reasoning.

This anatomically aware spatial bias offers domain grounding to the model's explanations and is particularly useful in ambiguous or complex radiographs where gradient-based or attention-based signals alone may lack spatial specificity.

\subsection{Multi-Component Attribution Fusion}

The final output of MATEX is a clinically grounded explanation map $A_{\text{MATEX}}$, computed via a weighted fusion of complementary information sources, as depicted in the fusion block (bottom-center) of Figure~\ref{fig:architecture}. This fusion balances raw gradient relevance, attention-based contextualization, anatomical localization, and layer stability filtering.

\textbf{Fusion Formula:}
\begin{equation}
A_{\text{MATEX}} = \alpha \cdot A_{\text{grad}} + \beta \cdot A_{\text{flow}} + \gamma \cdot (C \odot A_{\text{grad}}) + \delta \cdot (M \odot A_{\text{grad}})
\end{equation}

where the weights are configured as:
\begin{itemize}[leftmargin=*, noitemsep]
    \item $\alpha = 0.5$: Weight for gradient-based saliency, providing fine-grained importance
    \item $\beta = 0.2$: Weight for multi-scale attention rollout, reflecting semantic attention
    \item $\gamma = \lambda_c$: Configurable weight for consistency-modulated gradients (typically $0.3$--$0.4$)
    \item $\delta = 0.2$: Weight for anatomically-guided gradients
\end{itemize}

The term $(C \odot A_{\text{grad}})$ suppresses unreliable gradient regions with high inter-layer variability. The term $(M \odot A_{\text{grad}})$ reinforces gradient values that overlap with relevant anatomical regions inferred from the report.

\textbf{Weight Configuration:}
Unlike traditional normalized weighting schemes, MATEX uses a flexible weighting approach:
\begin{equation}
\alpha + \beta + \delta = 0.9 \quad \text{(fixed components)}
\end{equation}
\begin{equation}
\gamma = \lambda_c \quad \text{(configurable consistency weight)}
\end{equation}

This design maintains stable performance for base components while allowing practitioners to adjust the emphasis on layer consistency ($\gamma$) based on the reliability requirements of the specific clinical task, without requiring rebalancing of all other components.

\textbf{Interpretability Role.} This fusion enables \textit{clinical flexibility and control}—the user (radiologist or model developer) can emphasize the modality that aligns best with the diagnostic objective. For example:
- Emphasizing $A_{\text{flow}}$ (increasing $\beta$) enhances interpretability through transformer semantics.
- Increasing $\delta$ favors anatomically-aligned explanations.
- Boosting $\gamma$ enhances attribution robustness by filtering noisy gradients.

\section{Experiments and Results}

\subsection{Dataset and Implementation Details}

We evaluate \textit{MATEX} on the MS-CXR benchmark~\cite{boecking2022making}, a large-scale chest X-ray dataset featuring radiographic images paired with structured clinical reports containing spatially grounded annotations. The dataset includes localized pathologies such as consolidation, opacity, and pleural effusion, making it ideal for assessing spatial alignment and interpretability.

We use the pretrained ViT-B/32 CLIP encoder as the vision-language backbone. Gradient-based attribution maps are derived from the last 2–3 transformer layers, while multi-scale attention rollout aggregates attention across all layers to enhance spatial coverage and semantic richness.
Spatial priors are computed from anatomical phrases parsed from the reports, mapped to soft spatial masks. The spatial prior weight $\lambda_s$ and consistency weight $\lambda_c$ are tuned on a validation set within the ranges $\lambda_s \in [1.5, 3.5]$ and $\lambda_c \in [0.2, 0.6]$. All experiments are implemented in PyTorch and run on NVIDIA A100 GPUs.

\subsection{Evaluation Metrics}

To evaluate the effectiveness of \textit{MATEX}, we perform both quantitative and qualitative analyses.

\noindent \textbf{Quantitative Metrics.}
\begin{itemize}[leftmargin=*]
    \item \textit{Confidence Drop (↓):} Measures the reduction in model confidence after masking highly attributed pixels. Lower values suggest the highlighted regions are essential to the model's decision.
    \item \textit{Confidence Increase (↑):} Captures the rise in model confidence when minimally attributed pixels are removed. Higher values indicate more effective identification of irrelevant features.
    \item \textit{ROAR+ (↑):} Evaluates attribution quality by retraining the model after masking top-attributed regions. Higher scores imply stronger alignment between predictions and informative input areas.
    \item \textit{Localization Accuracy (↑):} Quantifies the spatial overlap between attribution maps and ground-truth pathology regions. Higher values reflect improved anatomical fidelity and alignment with clinical annotations.
\end{itemize}

\noindent \textbf{Qualitative Assessment.}  
In addition to quantitative metrics, we examine representative attribution maps to evaluate interpretability and clinical plausibility. This includes assessing whether highlighted regions correspond to anatomically meaningful structures and align with textual descriptors from associated radiology reports (e.g., “right upper lobe nodule”). Visual clarity, spatial precision, and consistency with expected pathology patterns are reviewed to determine how well attribution outputs support clinician reasoning and explainability in real-world diagnostic settings.

\subsection{Quantitative Results}

Tables~\ref{tab:quantitative_results_image_transposed} and~\ref{tab:quantitative_results_text_transposed} present quantitative evaluations on the MS-CXR dataset using both image-based and text-based metrics. We report three commonly used attribution quality indicators: confidence drop (lower is better), confidence increase (higher is better), and ROAR+ (higher is better), assessing both image and textual modalities.

\paragraph{Findings.}
MATEX achieves the best confidence increase (47.55\%) and lowest drop (0.51\%) in the image modality, indicating robust, decision-aligned attributions that surpass all baselines, including M2IB. For image ROAR+, MATEX ranks second (37.9\%) just below M2IB (38.7\%) but well ahead of methods like Saliency (25.46\%) and Chefer (24.42\%). In text-based evaluations, MATEX maintains competitive performance, with strong ROAR+ (16.98\%), low drop (1.91\%), and high gain (50.07\%). Although RISE achieves slightly better confidence increase (57.2\%) and lower drop (1.16\%), MATEX's higher ROAR+ (16.98\% vs 12.09\%) indicates more reliable attributions. Overall, MATEX consistently ranks at the top across both modalities, demonstrating its ability to deliver coherent, reliable explanations in vision-language medical models.

\begin{table}[htbp]
\centering
\caption{Quantitative evaluation on MS‑CXR – \textbf{Image metrics}. Best per column in bold.}
\label{tab:quantitative_results_image_transposed}
\begin{tabular}{lccc}
\toprule
\textbf{Method} & \textbf{Conf. Drop $\downarrow$} & \textbf{Conf. Incr. $\uparrow$} & \textbf{ROAR+ $\uparrow$} \\
\midrule
GradCAM~\cite{gradcam}        & 2.76 $\pm$ 0.03 & 12.64 $\pm$ 0.46 & 3.54 $\pm$ 0.80 \\
Saliency~\cite{saliency}      & 0.81 $\pm$ 0.01 & 35.08 $\pm$ 0.44 & 25.46 $\pm$ 1.35 \\
KS~\cite{ks}                  & 2.37 $\pm$ 0.04 & 10.24 $\pm$ 0.68 & 12.67 $\pm$ 1.02 \\
RISE~\cite{rise}              & 3.94 $\pm$ 0.03 & 7.28 $\pm$ 0.44 & 16.79 $\pm$ 0.76 \\
Chefer et al.~\cite{chefer}   & 1.87 $\pm$ 0.02 & 21.44 $\pm$ 0.46 & 24.42 $\pm$ 1.19 \\
M2IB~\cite{m2ib}              & 0.55 $\pm$ 0.01 & 45.92 $\pm$ 0.70 & \textbf{38.7 $\pm$ 0.86} \\
\textbf{MATEX (Ours)}         & \textbf{0.51 $\pm$ 0.03} & \textbf{47.55 $\pm$ 1.10} & 37.9 $\pm$ 0.79 \\
\bottomrule
\end{tabular}
\end{table}


\begin{table}[htbp]
\centering
\caption{MS‑CXR Text Metrics. Best results per column in bold.}
\label{tab:quantitative_results_text_transposed}
\begin{tabular}{lccc}
\toprule
\textbf{Method} & \textbf{Conf. Drop $\downarrow$} & \textbf{Conf. Incr. $\uparrow$} & \textbf{ROAR+ $\uparrow$} \\
\midrule
GradCAM~\cite{gradcam}        & 2.26 $\pm$ 0.04 & 36.24 $\pm$ 0.54 & 11.07 $\pm$ 0.62 \\
Saliency~\cite{saliency}      & 3.35 $\pm$ 0.03 & 18.88 $\pm$ 0.54 & 15.79 $\pm$ 0.92 \\
KS~\cite{ks}                  & 2.40 $\pm$ 0.05 & 34.12 $\pm$ 0.77 & 14.28 $\pm$ 1.09 \\
RISE~\cite{rise}              & \textbf{1.16 $\pm$ 0.02} & \textbf{57.2 $\pm$ 0.65} & 12.09 $\pm$ 1.52 \\
Chefer et al.~\cite{chefer}   & 2.93 $\pm$ 0.03 & 28.08 $\pm$ 0.34 & 9.11 $\pm$ 0.60 \\
M2IB~\cite{m2ib}              & 2.28 $\pm$ 0.04 & 35.48 $\pm$ 0.69 & 16.31 $\pm$ 0.75 \\
\textbf{MATEX (Ours)}         & 1.91 $\pm$ 0.06 & 50.07 $\pm$ 0.46 & \textbf{16.98 $\pm$ 0.66} \\
\bottomrule
\end{tabular}
\end{table}



\subsection{Qualitative Evaluation}

\begin{figure}[ht!]
\centering
\begin{tabular}{@{}c@{}c@{}}
\rotatebox{90}{\textbf{MATEX}} & 
\includegraphics[width=0.95\textwidth]{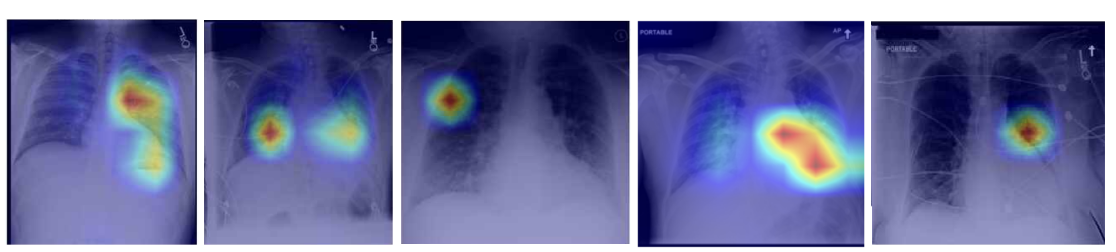} \\
\rotatebox{90}{\textbf{M2IB}} & 
\includegraphics[width=0.95\textwidth]{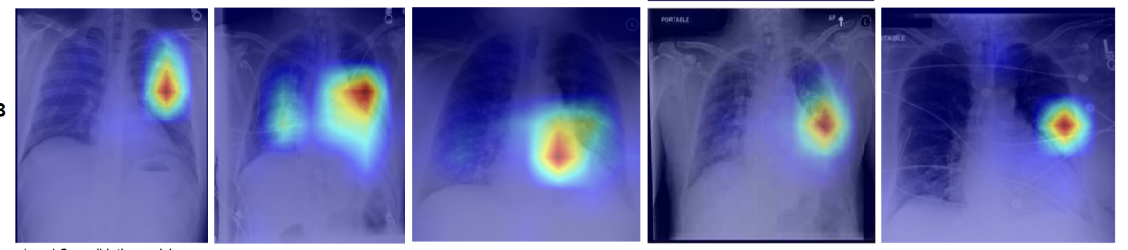} \\
\end{tabular}

\caption{
\textbf{Qualitative comparison between MATEX and M2IB on MS-CXR.}
Each column shows a clinical case with \textbf{MATEX (Top)} and \textbf{M2IB (Bottom)}.
\textbf{Left to Right:}
(1) \textit{Left upper lobe consolidation}: MATEX localizes sharply; M2IB is diffuse.
(2) \textit{Bilateral lower lobe consolidation}: MATEX captures both lobes; M2IB is less focused.
(3) \textit{Right mid-to-upper zone opacity}: MATEX aligns closely with pathology.
(4) \textit{Left lower lobe opacification}: MATEX offers better boundary definition.
(5) \textit{Left mid-lung consolidation}: MATEX maintains anatomical fidelity.
Overall, MATEX shows superior spatial precision and clinical alignment.
}
\label{fig:comparison}
\end{figure}

\noindent Figure~\ref{fig:comparison} shows a qualitative comparison between \textit{MATEX} and \textit{M2IB} on the MS-CXR dataset. Each column depicts a clinical case, with \textit{MATEX} heatmaps on the top row and \textit{M2IB} results below, visualized as overlayed attribution maps on chest X-rays.

\paragraph{Findings.}
Figure~\ref{fig:comparison} illustrates that MATEX offers several consistent improvements over M2IB. Across all evaluated cases, MATEX produces heatmaps with sharper localization and tighter alignment to pathological regions, enhancing anatomical coherence. In scenarios involving bilateral or multi-zone findings—such as lower lobe consolidation—MATEX accurately captures both affected zones while avoiding overspill, whereas M2IB often yields diffuse attention. Additionally, MATEX exhibits improved boundary adherence, particularly in cases of opacification, which enhances interpretability for clinical use. It also demonstrates reduced noise activation by limiting attribution to relevant structures, unlike M2IB, which sometimes highlights unrelated areas.

These improvements are attributable to MATEX’s ability to produce more anatomically precise and clinically aligned explanations. The framework effectively captures multi-region pathologies without spatial drift and maintains clear boundaries around relevant findings.


\section{Discussion and Limitations}

\noindent \textbf{Performance Summary.}
\textit{MATEX} improves interpretability in medical vision-language models by combining multi-scale gradient fusion, anatomical priors, and text-guided spatial consistency. Quantitative (Tables~\ref{tab:quantitative_results_image_transposed},~\ref{tab:quantitative_results_text_transposed}) and qualitative (Figure~\ref{fig:comparison}) results show that MATEX outperforms prior methods, achieving higher confidence increase and ROAR+ with minimal confidence drop. It also yields finer-grained localization and stronger anatomical alignment than M2IB, supporting its clinical relevance where spatial precision and trust are vital.

\noindent \textbf{Key Strengths.}
MATEX enables anatomically grounded attributions through spatial priors, captures both low- and high-level features via multi-scale attention, and fuses gradients with attention for rich explanations. Cross-modal fidelity keeps visual outputs aligned with clinical semantics.

\noindent \textbf{Limitations.}
MATEX adds $\sim$40\% overhead from multi-layer backpropagation. Its priors are currently specific to chest X-rays and require adaptation for other modalities. Attribution quality is sensitive to parameter tuning and the reliability of CLIP attention.

\section{Conclusion}

We introduced \textit{MATEX}, a framework that enhances interpretability in medical vision-language models by integrating multi-scale attention, anatomically-informed spatial priors, and consistency-driven refinement. By combining gradient signals with layer-wise attention rollout and clinical spatial cues, MATEX produces high-fidelity, anatomically grounded explanations for CLIP-based models. Experiments on the MS-CXR dataset show that MATEX outperforms GradCAM, RISE, and M2IB across both image- and text-based attribution metrics, with qualitative results confirming improved spatial precision and diagnostic relevance. These results demonstrate that incorporating textual semantics and spatial priors not only improves explanation quality but also aligns model outputs with clinical reasoning. While maintaining computational efficiency, MATEX addresses limitations in gradient-based methods such as spatial incoherence and modality misalignment.

\bibliographystyle{plain}
\bibliography{References}

\end{document}